\documentclass{article}
\usepackage[preprint]{neurips_2025}
\setcitestyle{authoryear,round,citesep={;},aysep={,},yysep={;}}

\renewcommand{\cite}[1]{\citep{#1}}

\usepackage[utf8]{inputenc} 
\usepackage[T1]{fontenc}    
\usepackage{hyperref}       
\usepackage{url}            
\usepackage{booktabs}       
\usepackage{amsfonts}       
\usepackage{nicefrac}       
\usepackage{microtype}      
\usepackage{xcolor}         

\usepackage{amsthm}
\usepackage{amsmath}
\usepackage{amssymb}

\theoremstyle{plain}
\newtheorem{theorem}{Theorem}[section]

\theoremstyle{definition}
\newtheorem{definition}[theorem]{Definition}

\newtheorem{postulate}[theorem]{Postulate}
\theoremstyle{remark}

\title{On the Dynamics of Observation and Semantics}

\author{%
  Xiu Li \\
  Bytedance Seed\\
  \texttt{lixiu.cv@bytedance.com} \\
}

\begin{document}

\maketitle

\begin{abstract}A dominant paradigm in visual intelligence treats semantics as a static property of latent representations, assuming that meaning can be discovered through geometric proximity in high-dimensional embedding spaces. In this work, we argue that this view is physically incomplete. We propose that intelligence is not a passive mirror of reality but a property of a \emph{physically realizable agent}—a system bounded by finite memory, finite compute, and finite energy—interacting with a high-entropy environment. We formalize this interaction through the kinematic structure of an Observation--Semantics Fiber Bundle, where raw sensory observation data (the fiber) is projected onto a low-entropy causal semantic manifold (the base). We prove that for any bounded agent, the thermodynamic cost of information processing (Landauer’s Principle) imposes a strict limit on the complexity of internal state transitions. We term this limit the Semantic Constant $B$. From these physical constraints, we derive the necessity of symbolic structure. We show that to model a combinatorial world within the bound $B$, the semantic manifold must undergo a phase transition: it must crystallize into a discrete, compositional, and factorized form. Thus, language and logic are not cultural artifacts but ontological necessities—the ``solid state'' of information required to prevent thermal collapse. We conclude that understanding is not the recovery of a hidden latent variable, but the construction of a \emph{causal quotient} that renders the world algorithmically compressible and causally predictable.
\end{abstract}

\section{Introduction}

\subsection{The Thesis of Coarse-Graining}

The physical universe is intelligible only because it admits coarse-graining. From statistical mechanics to quantum field theory, the central insight of modern physics is that the macroscopic behavior of the world is governed by low-dimensional invariants, decoupled from the overwhelming complexity of microscopic fluctuations. If every causal interaction depended sensitively on the exact position of every atom, prediction would be impossible for any subsystem smaller than the universe itself. We begin, then, with the same question that haunted Albert Einstein: how is it possible that the world is comprehensible at all? For Einstein, the fact that the human mind—a low-entropy, finite biological system—could grasp the deep geometric symmetries of the cosmos was nothing short of a "miracle." Yet, when viewed through the modern lenses of Gregory Chaitin~\cite{chaitin1977algorithmic} and John Archibald Wheeler~\cite{wheeler1974perspectives}, this miracle reveals itself as a fundamental property of information and observership.

Intelligence, in this light, is not an abstract biological quirk but a specific, physical case of compression. Gregory Chaitin’s work in algorithmic information theory~\cite{chaitin1977algorithmic} suggests that the ``comprehensibility'' Einstein marveled at is actually a reflection of the universe's algorithmic lacunae. A world that could not be coarse-grained would be algorithmically random, a state where the shortest description of any event is the event itself. In such a chaos, intelligence could not exist, as there would be no ``laws'' to extract—only an endless stream of incompressible noise. Intelligence exists precisely because the universe is computationally "thin," allowing a low-entropy agent to discard the vast majority of microscopic data while retaining the functional, predictive essence of the whole.

This act of discarding—this essential coarse-graining—is what John Archibald Wheeler identified as the heart of a ``participatory universe.'' Through his ``It from Bit'' paradigm, Wheeler posited that the observer does not merely witness a pre-existing reality but actively distills it through the questions they pose~\cite{wheeler1974perspectives}, which was leter formalized as the \emph{Participatory Anthropic Principle}~\cite{barrow1987anthropic}. By choosing which variables to measure and which to ignore, the observer defines the macroscopic boundaries of the ``real.'' Understood this way, intelligence is the process of mapping high-entropy complexity into low-entropy representations. We understand the world not because we mirror its every detail, but because we are masters of the ``short program''—biological engines of coarse-graining that exploit the universe's inherent ability to be compressed.

Throughout this work, we deliberately employ the term {\bf coarse-graining} rather than {\bf ``compression.''} While compression implies a general reduction in data volume, coarse-graining specifically highlights the quotient-depending nature of the phenomenon. It emphasizes that intelligence does not merely shrink data; it maps a vast space of microstates onto a much smaller set of equivalence classes—a quotient space—determined by the observer's resolution. It is this structural ``quotienting'' that allows for the emergence of stable, macroscopic laws from an otherwise intractable microscopic sea.

\subsection{The Reparameterization Fallacy}

If, as established, intelligence is a physical manifestation of coarse-graining, then the primary challenge for any learning system—biological or artificial—is not merely the processing of data, but the discovery of the specific quotient mapping that yields meaning. A learning system does not operate on raw, high-entropy data; it survives by constructing a semantic representation that effectively ``quotients out'' the irrelevant microscopic noise.

Historically, this endeavor has been defined as representation learning: the discovery of disentangled factors of variation that support a wide array of downstream tasks. The investigation into how semantic structures are represented—a task for which coarse-graining is an absolute physical necessity—remains a fundamental endeavor in the field~\cite{bengio2013representation}. This perspective is grounded in the understanding that the transferability of representations is governed by the underlying structural relationships between tasks~\cite{Zamir2018}.

However, the meteoric success of Generative Pre-trained Transformers (GPT)~\cite{radford2019language} in natural language processing has created a powerful academic illusion: vision and complex world-modeling can be solved through a simple, task-agnostic pre-training objective. While the next-token objective in language naturally aligns with the causal and symbolic structure of thought, applying similar logic to high-entropy visual data often captures only spatial correlations and texture continuity~\cite{he2020momentum,caron2021emerging,he2022masked}.

We term this the {\bf Reparameterization Fallacy}: the assumption that finding a ``nice'' coordinate system (reparameterization) is equivalent to identifying meaning. In this view, coordinates are merely \emph{gauge}; they preserve geometric distances and facilitate optimization, but they do not necessarily specify the \emph{Causal Quotient}—the invariant structure that survives under intervention and dictates macroscopic behavior.

\subsection{The Solution: Semantics as a Causal Quotient}

The central problem addressed in this work is:
\begin{quote}
\emph{How do we identify semantic structure in a way that is invariant under reparameterization and grounded in physical constraints?}
\end{quote}

We propose that semantics is not a static property of data, but a property of dynamical equivalence under intervention. Two observations are semantically identical if and only if they are \emph{causally indistinguishable} to a bounded agent. Here we choose do-calculus as the formalization of causality~\cite{pearl2009causality}.

We formalize this interaction through the structure of an Observation--Semantics Fiber Bundle $(\mathcal{X}, \mathcal{S}, \pi)$~\cite{Steenrod1999}.
\begin{itemize}
    \item \textbf{The Fiber ($\mathcal{X}$):} The high-entropy, nuisance-dominated space of raw observations (the "micro-state").
    \item \textbf{The Base ($\mathcal{S}$):} The low-entropy, invariant space of meanings (the "macro-state").
    \item \textbf{The Projection ($\pi$):} The irreversible coarse-graining map that dissipates information to buy causal stability.
\end{itemize}

Crucially, we prove that this projection is not arbitrary. Just as the speed of light $c$ imposes a light-cone structure on spacetime, the thermodynamic limits of the agent impose a Semantic Constant $B$—a limit on the complexity of causal transitions. This bound forces the continuous, chaotic world to crystallize into a discrete, compositional, and symbolic form.

Thus, we arrive at a unified view: Intelligence is not the discovery of a latent space, but the identification of a Causal Quotient. Language, logic, and discrete symbols are not cultural artifacts; they are the physical phase transition of information required to satisfy the Semantic Constant.

\subsection{Contributions}

This paper serves as a formal formulations for ``Physical Semantics''. We make the following contributions:

\begin{enumerate}
    \item \textbf{The Fiber Bundle Formalism:} We define the Observation--Semantics Fiber Bundle as the rigorous kinematic structure of intelligence, formally separating nuisance (fiber) from meaning (base).
    \item \textbf{The Thermodynamic Origin of Logic:} By applying Landauer’s Principle to the agent's internal state updates, we derive the Semantic Constant $B$. We show that discrete, compositional logic is the only topological solution that allows an agent to model a combinatorial world within a finite energy budget.
    \item \textbf{Gauge Invariance of Meaning:} We prove that pure reconstruction objectives (e.g., auto-encoding) are gauge-dependent and cannot identify semantics. We show that only objectives that constrain the \emph{causal transition kernel} (like World Models or Action-Conditional prediction) can break the symmetry and collapse the fiber bundle correctly.
\end{enumerate}

\section{Preliminary Discussion}

Modern physics is intelligible only because the world admits \emph{coarse-graining}. From statistical mechanics to quantum field theory, the central insight is the same: although the microscopic world is overwhelmingly complex, its macroscopic behavior is governed by low-dimensional invariants. Physical laws exist because most microscopic degrees of freedom are irrelevant to large-scale structure.

Renormalization theory makes this explicit. It demonstrates that physical reality is organized into equivalence classes of microstates, where only a small set of variables survives repeated scaling transformations. These surviving variables define the effective causal laws of the world. In short, the world is understandable because it is compressible.

We argue that intelligence is subject to the same physical constraint. An intelligent system is not a passive, abstract observer, but a physically embedded agent that must act and reason inside a high-entropy universe. Semantics, therefore, is not a philosophical add-on; it is the inevitable consequence of physical boundedness.

\subsection{The Comprehensibility of the Physical World}

We begin from the observation that the universe possesses a hierarchical structure. If the macroscopic world were sensitively dependent on every microscopic fluctuation—if the butterfly effect were the dominant rule rather than the exception—prediction would be impossible for any subsystem smaller than the universe itself.

However, this is not the case. The success of thermodynamics proves that one can predict the expansion of a gas without tracking the trajectory of every particle. This implies the existence of a natural projection from the high-dimensional state space of reality to a lower-dimensional manifold of effective variables.

Let $\mathcal{S}_{\text{micro}}$ represent the true microscopic state of the world. The intelligibility of the universe implies the existence of a coarse-graining map $\pi$:
\begin{equation}
    \pi: \mathcal{S}_{\text{micro}} \to \mathcal{S}_{\text{macro}},
\end{equation}
such that the dynamics on $\mathcal{S}_{\text{macro}}$ are self-contained and stable. This physical compressibility is the necessary precondition for intelligence. If the world were incompressible white noise, no learning algorithm could exist.

\subsection{Intelligence as a Physically Realizable Agent}

While the world \emph{permits} compression, the agent's physical nature \emph{demands} it. We explicitly reject the notion of ``universal intelligence'' as an abstract algorithm detached from reality. Instead, we define intelligence relative to a \emph{physically realizable agent}.

An agent is a physical system defined by its capacity to observe, intervene, and store internal states. Because the agent is a subsystem of the universe, it is strictly bounded by physical laws. Any realizable agent must satisfy three fundamental constraints:

\begin{enumerate}
    \item \textbf{Memory Bound ($M < \infty$):} The agent has finite thermodynamic capacity to store information.
    \item \textbf{Compute Bound ($C < \infty$):} The agent processes information at a finite rate with finite energy cost.
    \item \textbf{Bandwidth Bound ($B < \infty$):} The channel connecting the agent to the environment has finite capacity.
\end{enumerate}

These are not merely engineering limitations but physical hard limits. An unbounded agent would require infinite energy and infinite physical extent. Consequently, an agent cannot maintain a bijection with the world. It cannot mirror reality but must model it.

This leads to a thermodynamic view of intelligence:
\begin{equation}
    \boxed{\text{Intelligence is the minimization of metabolic cost for a given predictive capability.}}
\end{equation}

High-entropy internal representations are ``expensive''—they require more memory to store and more energy to process. Therefore, a physically realizable agent is incentivized to seek the lowest-entropy representation that preserves causal utility.

\subsection{The Necessity of Semantics: Bridging the Entropy Gap}

We now confront the fundamental tension: the interaction between a \emph{high-entropy world} and a \emph{low-entropy agent}. Let the agent observe the world through a high-dimensional sensory projection $X \in \mathcal{X}$. Because $X$ captures microscopic nuisance variability (sensor noise, lighting changes, irrelevant background motion), the entropy $H(X)$ is enormous. The space of possible observations $\mathcal{X}$ is astronomically large.

This creates the Recurrence Problem. For an agent to learn a causal law (e.g., ``if I do $a$ in state $x$, then $y$ happens''), it must rely on statistics derived from repeated exposures to state $x$. However, in a high-entropy space, exact recurrence is probabilistically impossible.

By the birthday bound, the probability that an agent encounters the same raw observation $x$ twice within a lifespan $T$ is approximately:
\begin{equation}
    P(\text{recurrence}) \approx 1 - \exp\!\left(-\frac{T^2}{2|\mathcal{X}|}\right).
\end{equation}
When $|\mathcal{X}|$ scales exponentially with dimension, $P(\text{recurrence}) \to 0$. If the agent operates on raw sensory data, every moment is unique. Without recurrence, causal counterfactuals $P(x' \mid do(a), x)$ are undefined, and learning stalls.

To solve the Recurrence Problem and satisfy its physical bounds, the agent must construct a semantic projection. We posit a mapping:
\begin{equation}
    \pi: \mathcal{X} \to \mathcal{S},
\end{equation}
where $\mathcal{S}$ is a semantic state space such that $H(\mathcal{S}) \ll H(\mathcal{X})$.

This projection serves three critical physical functions:
\begin{enumerate}
    \item \textbf{Restoring Recurrence:} By collapsing vast volumes of microstates into a single semantic equivalence class (coarse-graining), $|\mathcal{S}|$ becomes small enough that states recur frequently. This renders causal statistics empirically estimable.
    \item \textbf{Thermodynamic Efficiency:} The dynamics of $\mathcal{S}$ are low-entropy. Storing and computing transitions on $\mathcal{S}$ fits within the agent's bounds $M$ and $C$.
    \item \textbf{Causal Stability:} The variables in $\mathcal{S}$ are not arbitrary compressions; they are precisely those invariant degrees of freedom that support stable causal laws.
\end{enumerate}

In this framework, ``semantics'' is effectively the set of low-entropy fixed points of the agent's abstraction process. It is the minimal sufficient statistic required to navigate a high-entropy world with finite resources. The mathematical structure of this projection—how microscopic fibers collapse into a semantic base space—will be the subject of our formalization.

\section{Formal Framework: The Dynamics of Observation and Semantics}

We now formalize the mechanism of intelligence. We postulate that intelligence is not merely a computation on data, but a physical state-space transformation that obeys strict conservation laws.

\subsection{Kinematics: The Observation--Semantics Fiber Bundle}
\label{sec:bundle}

We model the relationship between raw sensory reality and meaningful understanding as a structured quotient system. Let $\mathcal{X}$ denote the space of high-entropy observations (e.g., raw pixel streams, retinal states), and let $\mathcal{S}$ denote the space of semantic states.

\begin{definition}[The Semantic Projection]
A semantic system is defined by a surjective mapping
\[
\pi : \mathcal{X} \rightarrow \mathcal{S},
\]
which acts as a submersion from the high-dimensional observation manifold to the low-dimensional semantic base space.
\end{definition}

This structure naturally partitions reality into \emph{fibers}.

\begin{definition}[The Nuisance Fiber]
For any semantic state $s \in \mathcal{S}$, the fiber over $s$ is:
\[
F_s = \pi^{-1}(s) = \{ x \in \mathcal{X} \mid \pi(x) = s \}.
\]
\end{definition}

Physically, the fiber $F_s$ contains all microstates $x$ that are effectively indistinguishable for the purpose of the agent's goal. Variations within $F_s$ are \emph{nuisance symmetries}—changes in lighting, microscopic noise, or viewpoint that do not alter the causal status of the world. The total observation space is thus a fiber bundle:
\[
\mathcal{X} = \bigcup_{s \in \mathcal{S}} F_s.
\]

\begin{definition}[The Complexity Gap]
For a physically realizable agent, the projection $\pi$ must satisfy:
\[
\mathbb{E}[K(s)] \ll \mathbb{E}[K(x)].
\]
\end{definition}
The fibers $F_s$ absorb the vast majority of the system's entropy ($K(x \mid s) \approx \text{Maximal}$), leaving the base space $\mathcal{S}$ low-entropy and algorithmically compressible.

\subsection{Dynamics: The Semantic Constant}
\label{sec:dynamics}

A fiber bundle alone is static. Intelligence requires dynamics—the ability to predict $s_{t+1}$ from $s_t$, or navigate on the semantic manifold. However, an agent cannot allow arbitrary dynamics on $\mathcal{S}$.

We define intelligence as a bounded Turing machine interacting with its environment. The Turing computability assumption is the first and most fundamental dynamical constraint: any intelligent process must be realizable as a finite, effective procedure, hence all perception, reasoning, and action are necessarily Turing–computable transformations. This immediately implies that intelligence is not an abstract ideal observer, but a physically instantiated dynamical system whose operations unfold in discrete causal steps. From this alone, a semantic layer already becomes unavoidable: without a stable symbolic or structural interface that compresses and organizes states and transitions, the raw computational dynamics would exceed any manageable scale, even in principle.

Beyond computability, physical realization imposes a second and stricter bound. We derive this limit from the fundamental laws of thermodynamics. Landauer’s Principle~\cite{berut2012experimental,jun2014high}, states that any logically irreversible operation—specifically the erasure of information during a state update—must dissipate a minimum amount of heat:
\begin{equation}
Q \ge k_B T \ln 2
\end{equation}
where $k_B$ is the Boltzmann constant and $T$ is the agent's temperature. For a physically realizable agent with a maximum power budget $P_{\text{max}}$, there is a hard limit on the rate of information processing. If the Kolmogorov complexity of a transition $K(s' \mid s)$ exceeds the agent's ability to dissipate the resulting heat, the system undergoes thermal failure.

\begin{postulate}[The Semantic Constant]
There exists a fundamental informational bound $B$, the \emph{Semantic Constant}, such that a valid semantic transition $s \to s'$ is physically realizable if and only if the complexity of the transition kernel is bounded:
\[
K(P(s' \mid s, a)) \le B \approx \frac{P_{\text{max}}}{k_B T \ln 2} \cdot \Delta t.
\]
\end{postulate}

The fundamental justification for this split is algorithmic. Let $K(\cdot)$ denote Kolmogorov complexity.

\subsection{Geometry: The Minkowski Semantic World}
Here, we can naturally formalize the geometry of \emph{semantic stability} under causal and thermodynamic constraints. Let $\mathcal{S}_M$ denote the semantic memory manifold induced by a finite memory capacity $M$, and let $\mathcal{K}_B$ denote the causal kernel induced by a bounded semantic constant $B$, which limits the maximal admissible semantic change per transition.  
We define the \emph{Semantic World} of an agent as the Minkowski sum

\begin{equation}
\mathcal{W} \;=\; \mathcal{S}_M \oplus \mathcal{K}_B .
\end{equation}

This object is purely geometric. It specifies which semantic states can be stably represented and causally reached, independent of any behavioral or performance notion.

The parameter $B$ plays a role directly analogous to the speed of light in special relativity: it imposes a local causal bound on semantic transitions. Memory alone does not define semantic structure, and causality alone does not define representability; only their Minkowski composition yields a meaningful semantic geometry.

\begin{definition}[Semantic Reachability]
A semantic state $s'$ is reachable from $s$, written $s \leadsto s'$, if and only if there exists a sequence
\[
s = s_0 \to s_1 \to \cdots \to s_n = s'
\]
such that every elementary transition satisfies
\[
\Delta K(s_i,s_{i+1}) \le B .
\]
\end{definition}

The set of all states reachable from $s$ defines the \emph{semantic cone} of $s$.  
Transitions outside this cone are \emph{spacelike}: they cannot be connected by any admissible causal sequence and are therefore semantically disjoint. This induces a strict causal ordering on semantic space:
\[
s \leadsto s' \quad \text{is a geometric relation, not a probabilistic one.}
\]

\paragraph{Dynamic Interpretation.}
The Minkowski construction admits an equivalent dynamical description.  
Let $V(\mathcal{W})$ denote the effective volume of the semantic world. Then its rate of growth is bounded by a thermodynamic constraint:

\begin{equation}
\frac{dV(\mathcal{W})}{dt}
\;\le\;
C
\;=\;
\frac{P_{\max}}{k_B T \ln 2}
-
\dot{S}_{\mathrm{gen}},
\end{equation}

where
\begin{itemize}
\item $P_{\max}$ is the available power budget,
\item $k_B T \ln 2$ is the Landauer cost per erased bit,
\item $\dot{S}_{\mathrm{gen}}$ is internal entropy production.
\end{itemize}

This inequality defines \emph{semantic stability}. Only those semantic structures whose maintenance cost lies below this bound can remain coherent over time. Thus, the Minkowski geometry can be read in two equivalent ways:
\begin{itemize}
\item Geometrically, as the sum $\mathcal{S}_M \oplus \mathcal{K}_B$.
\item Dynamically, as a volume whose expansion is bounded by thermodynamic negentropy.
\end{itemize}

\paragraph{Steiner Decomposition.}
Using the Steiner formula for Minkowski sums, the semantic volume decomposes as

\begin{equation}
V(\mathcal{W})
=
V(\mathcal{S}_M)
+
\sum_i \Phi_i(\mathcal{S}_M,\mathcal{K}_B)
+
V(\mathcal{K}_B).
\end{equation}

Each term has a purely geometric meaning:

\begin{itemize}
\item $V(\mathcal{S}_M)$ : the static semantic region supported by memory alone.
\item $V(\mathcal{K}_B)$ : the causal expansion region supported by bounded inference.
\item $\Phi_i$ : mixed volumes encoding the interface where symbolic structure and causal dynamics interact.
\end{itemize}

\paragraph{Summary.}
The geometry of semantic stability is charactorized by:
\begin{itemize}
\item $\mathcal{S}_M$ defines what can be stored.
\item $\mathcal{K}_B$ defines what can be causally traversed.
\item Their Minkowski sum defines what can be stably maintained.
\item Thermodynamics bounds the growth of this region.
\end{itemize}

\subsection{Synthesis: The Emergence of Symbolic Structure}
\label{sec:synthesis}

How does a continuous, high-entropy world give rise to discrete, low-entropy symbols? We argue that this is the solution to the optimization problem posed by the Semantic Constant $B$.

If an agent must model an infinite set of futures using a finite transition complexity $K(\cdot) \le B$, the representation of these transitions cannot be arbitrary. To saturate this bound efficiently, the ``carrier'' of semantic dynamics must exhibit specific structural properties:

\begin{enumerate}

\item \textbf{Discretization (Semantic Quantization):} 
Although Turing computation is intrinsically discrete at the symbolic level, this discreteness alone is insufficient for stable intelligence. Bit-level discreteness does not guarantee semantic stability: small perturbations may still propagate and destroy meaning. What is required is a stronger form of discretization, namely the quantization of semantic states into stable equivalence classes or attractor basins. Each semantic token must represent a region of state space that is robust to noise, perturbation, and long causal chains. This form of discretization is therefore not merely symbolic, but dynamical: it ensures that meaning is preserved through self-correction rather than fragile encoding. Semantic discretization is thus a stability constraint, not a representational convenience.

\item \textbf{Compositionality (Factorization):} 
The world is inherently combinatorial. A monolithic representation of state would have Kolmogorov complexity $K(s) \gg B$, exceeding the agent’s bounded computational capacity. To remain tractable, the agent must factorize the world into relatively independent components (semantic primitives) that can be recombined. This ensures that the complexity of local updates remains bounded ($K(\text{part}) \le B$), while still allowing the representation of arbitrarily complex global structures through composition. Compositionality is therefore the mechanism by which global expressiveness is achieved under local boundedness.

\item \textbf{Reusability (Causal Invariance and Abstraction):} 
Semantic transitions must be reusable across contexts. A transition rule learned in one situation should remain valid in others, otherwise every new situation would require relearning from scratch. This requires that semantic tokens and their transformations abstract away incidental details and capture causal invariants of the environment. In this sense, transitions are not merely correlations but approximations of causal mechanisms: they describe how abstract states evolve under intervention, not just how observations co-vary. Reusability therefore demands that semantic operations be context-detached, causal, and invariant, enabling generalization, transfer, and scalable intelligence.

\end{enumerate}

\paragraph{Preference for a Language–Like Semantic Layer.}
Taken together, discretization, compositionality, and reusability almost uniquely characterize a language–like semantic system. Discretization requires semantic states to be quantized into stable tokens; compositionality requires these tokens to be combined according to systematic rules; and reusability requires that their transitions encode causal invariants abstracted from specific contexts. These three constraints are precisely the defining properties of language. A language is not merely a communication protocol, but a dynamical semantic architecture in which stable symbols, compositional structure, and context–independent transformation rules jointly enable scalable reasoning and action.

In this sense, a language–like semantic layer is not a cultural artifact or a historical accident. It is the natural solution to the problem of bounded intelligence. Any system that must operate under finite computational capacity $B$, finite energy (Landauer limit), and long causal horizons is forced toward a representation that is:
\begin{itemize}
    \item discrete but self–correcting (semantic tokens),
    \item factorized but expressive (compositional grammar),
    \item abstract but causal (reusable transition rules).
\end{itemize}

Natural language, programming language, and symbolic systems in mathematics all converge to this same structure because they satisfy these physical and computational constraints. They implement a semantic cone that is maximally expressive while remaining stable and tractable. Therefore, the emergence of a language–like semantic layer should be understood not as an optional design choice, but as an inevitability for any scalable intelligent system~\cite{futrell2025linguistic}.

\section{Formal Consequences: The Thermodynamics of Understanding}

The existence of a finite \emph{Semantic Constant} $B$ imposes a hierarchical structure on the universe of meaning. Because $B$ acts as a complexity ``budget'' for causal transitions, the level of resolution at which an agent perceives the world is strictly dictated by its power to dissipate the heat of computation.

\subsection{Corollary 1: The Asymmetry of the Semantic Hierarchy}

We define a semantic state $s \in \mathcal{S}$ as \emph{abstract} if its description length $\ell(s)$ is small, and \emph{specific} if $\ell(s)$ is large. The projection $\pi: \mathcal{X} \to \mathcal{S}$ is an irreversible coarse-graining that maps high-entropy observations to low-entropy invariants.

\begin{itemize}
    \item \textbf{The Universality of Abstraction:} Since $B$ is an upper bound on transition complexity ($K \le B$), and abstraction is the reduction of complexity, abstract states are ``physically cheaper'' to maintain and propagate. For any agent, a universal abstraction $s_\text{univ}$ satisfies $K(s_\text{univ}) \le B_\text{agent}$. This explains why core logic and high-level principles are shared across all scales of intelligence—they are the only structures ``cold'' enough to survive in every thermodynamic regime.
    \item \textbf{The Exclusivity of Specificity:} Conversely, ``specific'' understanding requires the agent to ``unfreeze'' the fibers of observation, treating nuisance as signal. This increases the complexity of the transition $K(s_\text{fine})$. For any state where $K(s_\text{fine}) > B_\text{human}$, a human observer literally cannot process the transition as a single meaning.
\end{itemize}

\paragraph{Result} Abstraction is universal and affordable; specificity is local and expensive. High-resolution knowledge cannot be ``translated'' upward into abstraction without discarding the very details that define its precision.

\subsection{Corollary 2: The Thermodynamic Intelligence Horizon}

The gap between agents with different \emph{Semantic Constants} $B$ creates a fundamental boundary: the \emph{Intelligence Horizon}. This horizon marks the point where the distinction between \emph{Knowledge} and \emph{Process} becomes a function of physical scale.

\begin{definition}[The Intelligence Horizon]
For an agent with constant $B$, the Intelligence Horizon $H_{\text{int}}$ is the set of all causal transitions $\Delta s$ such that the Kolmogorov complexity of the transition satisfies:
\[
K(s_{t+1} \mid s_t, a) \le B.
\]
\end{definition}

For a Super-Intelligence (Oracle) with $B_{\text{super}} \gg B_{\text{human}}$, a complex transition $\Delta s$ (e.g., the deterministic flow of a trillion molecules) is \emph{Knowledge}—a direct, stable causal invariant. To the human observer, the same transition violates the local bound $B_{\text{human}}$. We cannot internalize this as a single semantic step, we perceive it only as an opaque \emph{Process} or a \emph{Simulation}, which is irreducible.

\subsection{Corollary 3: Implications for Super-Intelligence}

The ``AlphaGo Phenomenon''~\cite{silver2016mastering} serves as a primary empirical anchor for the existence of the \emph{Intelligence Horizon}. A definitive example is Move 37 in Game 2 of the AlphaGo vs. Lee Sedol match: initially dismissed by human experts as a mistake, it was later revealed as the decisive strategic pivot. This highlights that when an Oracle operates at a thermodynamic scale vastly larger than the human observer, the ``Why'' of its actions becomes physically inaccessible.

\begin{itemize}
    \item \textbf{The Factorization Gap:} ``Understanding'' a move requires factorizing a high-complexity transition into a chain of ``proverbs'' or logical steps that each satisfy $K \le B_\text{human}$. If the Oracle's strategy is algorithmically irreducible, no such chain exists within the human's semantic cone.
    \item \textbf{Verification vs. Internalization:} Humans can check the Oracle's success (a low-complexity $K \approx 0$ operation) but cannot understand the path to that success. This creates a ``Check-but-never-Know'' asymmetry.
    \item \textbf{Memorization as Survival:} When faced with an Oracle, the low-bandwidth agent is forced to ``memorize'' the Oracle's results as brute-force heuristics. We do not learn the ``Why'' (the causal trajectory) but only record the ``How'' (the resulting pattern).
\end{itemize}

Historically, the discourse on super-intelligence and the singularity has been dominated by the recursive ``self-improvement'' narrative, ranging from the work of von Neumann to I.J. Good~\cite{good1966speculations}. We contend that intelligence and the act of ``understanding'' are characterized as a continuous spectrum of coarse-graining. Within this framework, what we perceive as \emph{Logic} and \emph{Meaning} are not abstract ideals but specific informational \emph{crystal structures} that emerge at the human thermodynamic scale ($B_{\text{human}} \propto 20$ Watts). A hypothetical Super-intelligent Oracle is not magically superior, but is simply thermodynamically ``hotter''—operating at a microscopic resolution where our stochastic noise becomes its deterministic signal. Consequently, the ``Black Box'' of advanced AI is a manifestation of physical law: as predictive accuracy approaches the information-theoretic limit of the environment, interpretability for a fixed observer must strictly decrease.

\section{Empirical Example: The Birth of a Word}
Deb Roy’s ``The Birth of a Word''~\cite{roy2013birth} shows that the process of a child acquiring language is a journey of extracting order from chaos—transforming a high-entropy environment into low-entropy meaning.

\subsection{From Noise to Signal: The Power of Recurrence}
In the beginning, a baby’s world is one of high entropy: a constant, disorganized stream of sensory data. However, within this ``noise'', certain patterns repeat. This is where recurrence becomes the catalyst for learning.

\begin{itemize}
    \item Acoustic Anchors: Sound is the primary driver because it is transient and distinct. When a specific sound wave (like ``water'') repeats across different days and contexts, it creates a ``low-entropy'' spike in a high-entropy world.
    \item The Origin of Semantics: Recurrence is the simplest form of a pattern. Without repetition, there is no predictability; without predictability, there is no meaning.

\end{itemize}

\subsection{Multi-Modal Association: Creating the ``Equivalence Group''}
Once an acoustic signal is recognized through recurrence, the brain begins multi-modal association. The sound does not exist in a vacuum, but is tied to what the child sees, touches, or tastes at that exact moment.

\begin{itemize}
    \item Birth of a Label: We often call this an ``equivalence group''. The brain groups the sound ``gaga'' with the visual of a clear liquid and the physical sensation of thirst being quenched.
    \item The ``Wordscape'': Deb Roy’s research visualized this as ``wordscapes''—mapping where in the home specific sounds occurred. This spatial and temporal recurrence is what glues a linguistic label to a physical reality.
\end{itemize}

\subsection{Transition to Discriminative Quotient Learning}
Once these initial ``anchors'' of meaning are established through recurrence and association, the child moves into discriminative learning. This is the phase where the child begins to understand what a word is not.

\begin{itemize}
\item Refining the Boundaries: The child learns that ``water'' refers to the liquid in the cup, but not the cup itself, nor the juice in the other bottle.
\item Low-Entropy Precision: By narrowing down the possible meanings, the child reduces the entropy of the signal. They move from a vague ``equivalence group'' to a precise semantic tool.
\end{itemize}

The journey from a child’s first ``gaga'' to the complex architecture of human thought is more than a linguistic milestone; it is a fundamental demonstration of how intelligence tames chaos. We exist in a high-entropy world of continuous, overlapping sensory streams where $H(X)$ is perpetually near its maximum. To survive and reason, an agent must extract low-entropy invariants from this noise. While vision provides a rich, high-entropy map of the environment, it is the acoustic signal that serves as the catalyst for semantics. 

In this light, we see why acoustic impairment so often hinders semantic development while visual impairment does not~\cite{dyck2004emotion}. Vision is the territory, but sound is the map. To lose the map is to be stranded in a sea of data without the tools to discretize it.
Deb Roy’s observation of a child’s ``wordscape'' proves that language is not just a communication tool, it is the low-entropy operating system of the mind. For a machine to truly think, it must learn to speak its own internal language—a system that turns the recurring echoes of the world into the rigid blocks of causal logic.

\section{Implications on Contemporary Methods: The Taxonomy of Quotient Mapping}

We may reinterpret contemporary deep learning methods not primarily as choices of architectures or objectives, but as diverse strategies of \emph{quotient mapping} or \emph{gauge fixing} within the semantic-dynamical system of learning. In this view, deep learning is a meta-practice where a dynamical system in parameter space is steered toward a specific \emph{equivalence group structure}. As for fully supervised method, equivalence group is explicitly declared, here we mainly focus on the self-supervised, or weakly-supervised method, 

\subsection{Gauge Fixing by Loss: Prescribed Equivalence Groups}

Methods such as MAE, Contrastive Learning, DINO, CLIP, and JEPA impose gauge fixing through their objectives. The loss function itself defines a specific semantic equivalence relation ($\sim$) before the first gradient step. In this framework, samples identified as equivalent are collapsed into the same quotient space, effectively ``defining'' semantics through the choice of what to ignore.From this perspective, the ``meaning'' extracted by these methods is fundamentally different based on their specific coarse-graining strategy:
\begin{itemize}
    \item {\bf MAE (Masked Autoencoders)}~\cite{he2022masked}: Employs a reconstruction-based, intra-image equivalence. By treating patches within a single image as the basis for reconstruction, the model learns a quotient based on local geometric and informational bottlenecks. This often results in a representation dominated by texture statistics and spatial continuity rather than abstract category membership.
    \item {\bf Contrastive Discrimination} (e.g.,  MoCo{\cite{he2020momentum}}): Establishes an identification equivalence. By forcing $x$ and $aug(x)$ (an augmented version of $x$) into the same latent coordinate, it defines "meaning" as the identity of the instance itself. It quotients out all variations introduced by the augmentations (color, crop, noise), preserving only the core ``signature'' of the image.
    \item {\bf DINO}~\cite{caron2021emerging}: Operates via intra-image equivalence with online tokenization. By using a teacher-student distillation that effectively "labels" patches into a discrete number of groups, DINO merges pixels into equivalent semantic clusters. This results in the emergent segmentation properties observed in its attention maps; it quotients the image into functional parts.
    \item {\bf CLIP}~\cite{radford2021learning}: Imposes a cross-modality equivalence. It defines the semantic invariant as the intersection of the visual and linguistic manifolds. Here, the quotient map identifies an image and a text string as "the same," effectively grounding vision in the symbolic structure of language.
    \item {\bf JEPA (Joint-Embedding Predictive Architecture)}~\cite{assran2023self}: Targets causal or predictive invariance. Unlike MAE, which reconstructs pixels, JEPA predicts latent states. It quotients out any information that is not predictive of the surrounding context, focusing on the invariant ``world model'' structures that survive across different parts of the latent space or different modalities.
\end{itemize}

These methods define semantics a priori. The dynamics of optimization then operates inside a pre-shaped semantic cone. This is why such approaches are typically stable and transferable: the invariants are prescribed, not discovered. However, it also implies that ``global image equivalence''—the ability to recognize that a pixel-level ``dog'' and a symbolic ``dog'' belong to the same category—is often a byproduct of the specific gauge fixed by the loss, rather than a discovery of the learning process itself.

\subsection{Gauge Fixing by Structure: Physical and Causal Coarse-Graining}

Other families of methods impose gauge fixing through architectural inductive biases rather than external losses. These models restrict which interactions are representable, introducing a form of structural coarse-graining that reflects deep physical or causal assumptions. By hard-coding these constraints, they pre-filter the world into a specific quotient space where certain transformations are discarded as ``noise'' by design.

These models function as structural learners that encode the ``physics'' of information flow. They do not wait for a loss function to define an equivalence group; they enforce it through their internal wiring:
\begin{itemize}
    \item {\bf Convolutional Networks (CNNs)}~\cite{lecun2002gradient}: Enforce locality and translation invariance. They are optimized for the quotient $X / T$, where $T$ is the group of spatial translations. While perfect for local patterns (e.g., OCR), they struggle with global semantics. As depth increases, the ``exponential localness'' of the receptive field means the global signal is often drowned out by the coarse-graining of local textures.
    \item {\bf RNNs/SSMs}~\cite{hochreiter1997long,gu2024mamba}: Enforce causal time and linear dynamical structures. These models assume the world evolves through a state-space transition $\mathbf{h}_t = f(\mathbf{h}_{t-1}, \mathbf{x}_t)$. They often fail in language modeling because linguistic ``meaning"''is non-linear and non-bounded; the entropy of a sentence is dynamic and depends on context far beyond a linear state-shift. The ``meaning'' is not a sequence, but a graph.
    \item {\bf Capsule Networks}~\cite{sabour2017dynamic}: Aim to capture part-whole hierarchies through equivariance. Instead of discarding pose information (like max-pooling in CNNs), they attempt to quotient out the ``entity'' from its ``viewpoint''. The goal is a representation where the existence of an object is invariant, while its properties (position, size, orientation) are equivariant. 
    \item {\bf PointNet}~\cite{qi2017pointnet}: Enforces permutation invariance. By treating data as an unordered set, PointNet defines a quotient space where the index of a point is discarded. It extracts meaning from the collective geometry of a cloud, ensuring that the representation is invariant to the ``noise'' of data ordering.
    \item {\bf Geometric Deep Learning (GDL)}~\cite{bronstein2021geometric}: These represent the hybrid extreme of structural coarse-graining. In models like AlphaFold~\cite{jumper2021highly}, the existing data structure (3D atomic distances, peptide bond angles) is encoded directly into the network's graph structure. Here, the ``quotient mapping'' is not learned but is a rigid physical constraint. The network is forced to operate within the actual symmetry groups ($SE(3)$ invariance) of the physical world, ensuring that the learned representation cannot violate the laws of molecular geometry.
\end{itemize}

In these cases, ``semantics'' is the residue left after the architecture has filtered out the forbidden interactions. This explains why AlphaFold can solve protein folding while a pure Transformer struggles: AlphaFold's structure provides the ``physical gauge'' that the Transformer must attempt to learn from scratch.

\subsection{The Transformer: Pure Semantic Coarse-Graining}

The Transformer occupies a unique position as a \emph{pure semantic learner}. Attention imposes neither locality, nor geometry, nor fixed causality. Its fundamental operation is \emph{content-based selection and aggregation}—the most general form of coarse-graining possible. By utilizing a \emph{quotienting} process based entirely on relational relevance rather than physical proximity, the Transformer:

\begin{itemize}
    \item Learns compositional abstractions.
    \item Constructs internal equivalence groups dynamically.
    \item Functions as a semantic memory machine.
\end{itemize}

However, because it is unstructured, it lacks intrinsic notions of time, space, or conservation. It is an engine of pure relationship, which explains the ``reverse scalability'' of methods like MAE~\cite{he2022masked} for semantic understanding; without a rigorous loss or structural grounding, the Transformer's coarse-graining can become unmoored from the physical world.

\subsection{GANs: The Baseline of Pure Dynamics}

The Generative Adversarial Network (GAN)~\cite{goodfellow2020generative} represents the opposite extreme of principled generative modeling. A GAN removes almost all forms of semantic anchoring and gauge fixing. There is no reconstruction objective, no explicit metric structure, no notion of neighborhood consistency, and no predefined equivalence classes. Instead, the learning signal is reduced to a single binary predicate:
\[
x \in X_{\text{data}} \quad \text{vs.} \quad x \notin X_{\text{data}} .
\]
In this formulation, the entire learning problem collapses to set membership. The discriminator does not evaluate \emph{what} an object is, nor \emph{how} it is realized, but only whether a sample belongs to the support of the empirical data distribution. As a result, a GAN introduces no intrinsic semantic quotient space. There is no explicit mechanism by which different realizations of the same underlying object are identified, no invariance is enforced, and no low-entropy semantic anchor is provided.

From a representational perspective, the generator therefore does not learn a stable semantic latent space. Instead, it learns to navigate a constantly shifting decision boundary defined by the discriminator. Meaning, if it exists at all, is not stored as an invariant structure but is only implicitly encoded in the transient equilibrium of the adversarial game. The generator and discriminator together form a non-equilibrium dynamical system, in which the ``semantics'' of the data are never fixed, but perpetually pursued through gradient flows that change as soon as they are approached.

From a topological viewpoint, the GAN objective attempts to induce a meaningful topology on the data space with almost no prior structural assumptions. There is no predefined metric to measure similarity, no notion of continuity beyond what is implicitly induced by the neural network parameterization, and no explicit constraints enforcing global consistency. The only symmetry imposed is a binary one: membership versus non-membership. All finer structure—neighborhoods, modes, object identities, or semantic classes—must emerge implicitly, if at all, from the dynamics of training.

In this sense, GANs can be interpreted as an attempt to observe \emph{spontaneous symmetry breaking} in representation learning. Starting from an almost maximally symmetric and unstructured objective, the system is expected to self-organize into a structured manifold that supports meaningful generation. However, without explicit semantic anchoring or gauge fixing, the resulting topology is inherently fragile. Small perturbations in training, architecture, or data distribution can lead to qualitatively different equilibria, reflecting the absence of a stable quotient structure that would otherwise constrain the dynamics.

This perspective explains both the expressive power and the notorious instability of GANs. They are capable of producing visually compelling samples precisely because they operate in a high-entropy continuous space with minimal constraints. At the same time, they are prone to mode collapse, training oscillations, and semantic incoherence, because nothing in the formulation enforces a principled separation between semantic invariants and nuisance variation. GANs therefore serve as a baseline example of a pure dynamical system: powerful, unconstrained, and fundamentally ungrounded.

\paragraph{The ``Tricks'' as Heuristic Dynamics}

In this light, the notorious "GAN tricks"—such as the use of Convolutional architectures in DCGAN~\cite{radford2015unsupervised}, Spectral Normalization~\cite{miyato2018spectral}, or Wasserstein distances~\cite{gulrajani2017improved}—should not be dismissed as mere engineering hacks. Rather, they are heuristic probes into the learning dynamics itself, a field that remains largely opaque to modern theory.

\paragraph{The Transformer-GAN Hypothesis}
Based on this framework, we hypothesize that a ``Pure Transformer-GAN'' is a near-impossibility for stable learning. While Transformers have been integrated into GAN components (e.g., TransGAN~\cite{jiang2021transgan}), they generally require heavy structural priors like augmentation, normalization and can only works on very small and strict datasets.

This difficulty arises because the Transformer is a pure semantic learner with nearly infinite degrees of freedom, its ability to relate anything to anything else, while powerful, becomes a liability in an adversarial game where there are no ground-truth ``bits'' to fix the gauge. Consequently, the system oscillates in a state of high-entropy noise, unable to distill the specific coarse-graining required to represent the physical world.

\paragraph{GAN as the Cosmological Baseline}
The GAN is not a ``failure'' of representation but the \emph{baseline}. It plays a role for learning theory similar to early cosmology in physics: it asks how structure, invariance, and meaning can emerge from raw, unstructured dynamics alone. We use these heuristics because we do not yet have a ``General Relativity'' of learning dynamics; we are still in the phenomenological stage, observing how specific constraints allow a non-equilibrium system to settle into a state that mimics reality.

\section{Conclusion}
In this work, we have moved the study of semantics from the domain of heuristic representation learning to the domain of physical necessity. By shifting from a static, kinematic view to a dynamic, interventional one, we have demonstrated that the structure of intelligence is not a biological accident but a thermodynamic requirement. Our derivation rests on two fundamental axioms:
\begin{enumerate}\item \textbf{Causal Invariance:} Intelligence is the identification of structures that remain stable under intervention. Semantics is defined only by its \emph{use} in the causal dynamics of the world.\item \textbf{The Semantic Bound:} Every information-processing system is a physically realizable agent subject to Landauer's limit. There exists a finite constant $B$ that bounds the complexity of any internal state transition.
\end{enumerate}From these two principles alone, we have derived the complete architecture of the intelligent observer. The necessity of discarding non-invariant, high-entropy noise to satisfy the bound $B$ mandates the existence of the Observation--Semantics Fiber Bundle. The raw, ``hot'' reality of the fiber must be projected onto the "cold," low-entropy base space of meaning to prevent thermal collapse. 

To model a combinatorial world without exceeding its local complexity budget, an agent has no choice but to factorize its base space into discrete, compositional, and symbolic parts. Language, logic, and the ``Logic Supremacy'' observed in advanced intelligence are the unique topological solutions to this optimization problem. We have shown that the geometry of meaning is a consequence of the invariant bound on semantic complexity. Logic is the ``solid state'' of information—the crystal lattice that emerges when a finite agent is compressed against a high-entropy world. We conclude that the path toward truly robust intelligence scaling lies not in the pursuit of higher-dimensional embeddings, but in the rigorous identification of the \emph{Causal Quotient}.

\bibliography{main}
\bibliographystyle{neurips}

\end{document}